\documentclass{article}

\usepackage{arxiv}

\usepackage[utf8]{inputenc} % allow utf-8 input
\usepackage[T1]{fontenc}    % use 8-bit T1 fonts
\usepackage{hyperref}       % hyperlinks
\usepackage{url}            % simple URL typesetting
\usepackage{booktabs}       % professional-quality tables
\usepackage{amsfonts}       % blackboard math symbols
\usepackage{nicefrac}       % compact symbols for 1/2, etc.
\usepackage{microtype}      % microtypography
\usepackage{cleveref}       % smart cross-referencing
\usepackage{lipsum}         % Can be removed after putting your text content
\usepackage{graphicx}
\usepackage{natbib}
\usepackage{doi}
\usepackage{multirow}
\usepackage{booktabs}
\usepackage{xcolor}

\title{OpenThaiGPT 1.6 and R1: Thai-Centric Open-Source and Reasoning Large Language Models}

\date{}

\newif\ifuniqueAffiliation
\ifuniqueAffiliation % Standard variant of author block
\author{{Sumeth Yuenyong\thanks{ORCID: 0000-0001-7774-0291}} \\
	Department of Computer Science,\\ Faculty of Engineering, Mahidol University, Thailand\\
	\texttt{sumeth.yue@mahidol.edu} \\
	\And
	{Thodsaporn Chay-intr\thanks{ORCID: 0000-0002-7287-7579}} \\
	iApp Technology Co., Ltd., Thailand\\
	\texttt{t.chayintr@gmail.com} \\
	\AND
        {Kobkrit Viriyayudhakorn\thanks{ORCID: 0000-0001-6700-7248}} \\
	Artificial Intelligence Entrepreneur Association of Thailand (AIEAT)\\
        iApp Technology Co., Ltd., Thailand\\
	\texttt{kobkrit@aieat.or.th} \\
}
\else
\usepackage{authblk}

\author[1]{%
	{Sumeth Yuenyong\thanks{\texttt{sumeth.yue@mahidol.edu}, ORCID: 0000-0001-7774-0291}}%
}
\author[2,3]{%
	{Thodsaporn Chay-intr\thanks{\texttt{t.chayintr@gmail.com}, ORCID: 0000-0002-7287-7579}}%
}
\author[2,3,4]{%
	{Kobkrit Viriyayudhakorn\thanks{\texttt{kobkrit@aieat.or.th}, ORCID: 0000-0001-6700-7248}}%
}
\affil[1]{Department of Computer Science, Faculty of Engineering, Mahidol University, Thailand}
\affil[2]{iApp Technology Co., Ltd., Thailand}
\affil[3]{Intelligent Informatics and Service Innovation Research Center, Thailand}
\affil[4]{Artificial Intelligence Entrepreneur Association of Thailand (AIEAT), Thailand}
\fi

\renewcommand{\headeright}{Technical Report}
\renewcommand{\undertitle}{\includegraphics[width=0.25\textwidth]{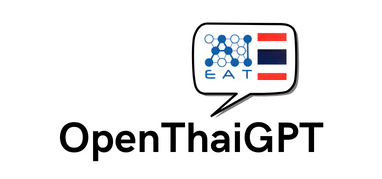}\\[0.5ex]Technical Report}
\renewcommand{\shorttitle}{OpenThaiGPT 1.6 and R1}

\hypersetup{
pdftitle={OpenThaiGPT 1.6 and R1: Thai-Centric Open-Source and Reasoning Large Language Models},
pdfsubject={cs.CL},
pdfauthor={Sumeth Yuenyong, Thodsaporn Chay-intr, Kobkrit Viriyayudhakorn},
pdfkeywords={LLM, Thai},
}

\begin{document}
\maketitle

\begin{abstract}
We present OpenThaiGPT 1.6 and R1 (OTG-1.6 and OTG-R1), Thai-centric Large Language Models (LLMs) developed through distinct methodologies to enhance generalization and reasoning capabilities. OTG-1.6 employs Task Arithmetic model merging for broad generalization, while OTG-R1 integrates multi-stage training with the Less-Is-More Reasoning Hypothesis (LIMO) for advanced reasoning. Benchmark evaluations demonstrate superior performance across Thai language tasks, achieving competitive results against larger-scale open-source Thai LLMs. This paper details the proposed models, training processes, benchmarks, and results, highlighting improvements over previous models and establishing new performance standards for Thai-centric LLMs.
\end{abstract}

\section{Introduction}

The development of Large Language Models (LLMs) has significantly advanced natural language understanding, reasoning, and generation capabilities \citep{openai2024openaio1card,gemmateam2025gemma3technicalreport,deepseek-2025-r1,xu2025largereasoningmodelssurvey}. Recent models have demonstrated impressive performance across various tasks by leveraging sophisticated architectures, extensive training data, and improved training methodologies \citep{qin2024largelanguagemodelsmeet,luo2025largelanguagemodelagent,openai2025competitiveprogramminglargereasoning}. However, achieving optimal performance for Thai-centric LLMs remains challenging due to linguistic complexities, limited high-quality datasets, and inadequate adaptation of general LLM architectures to Thai-specific tasks.

Addressing these challenges requires techniques that enhance generalization, reasoning, and efficiency without excessively increasing model scale. We present \textbf{OpenThaiGPT 1.6 (OTG-1.6)} and \textbf{OpenThaiGPT R1 (OTG-R1)}, developed using complementary methodologies to overcome these limitations. \textbf{OTG-1.6} applies Task Arithmetic model merging to combine specialized models, improving generalization across diverse tasks without increasing computational requirements \citep{ilharco-2023-editing,goddard-etal-2024-arcees}. Meanwhile, \textbf{OTG-R1}, inspired by DeepSeek-R1 \citep{deepseek-2025-r1}, employs multi-stage training and the Less-Is-More Reasoning Hypothesis (LIMO) to enhance reasoning capabilities with limited data \citep{ye-2025-limo-reasoning}.

Benchmark evaluations demonstrate that OTG-1.6 and OTG-R1 achieve superior performance across various Thai language tasks, with OTG-1.6 excelling in generalization tasks and OTG-R1 achieving competitive or superior results on reasoning benchmarks despite its smaller model size. 

The following sections present the proposed models, training processes, evaluation benchmarks, and results, highlighting their improvements over existing Thai-centric LLMs \citep{pipatanakul2023typhoonthailargelanguage,pipatanakul2024typhoon2familyopen,PathummaText}.

\section{Approach}

The development of OTG-1.6 and OTG-R1 involves the training of two distinct models: OTG-1.6, a general-purpose model with 72B parameters, and OTG-R1, a reasoning-enhanced model with 32B parameters. Both models were trained using specialized datasets and fine-tuning techniques aimed at improving generalization and reasoning. The base model for OTG-1.6 is \texttt{Qwen2.5-72B-Instruct}\footnote{\url{https://huggingface.co/Qwen/Qwen2.5-72B-Instruct}}, while the base model for OTG-R1 is \texttt{DeepSeek-R1-Distill-Qwen-32B}\footnote{\url{https://huggingface.co/deepseek-ai/DeepSeek-R1-Distill-Qwen-32B}}. Training was conducted using the MS Swift framework \citep{zhao2024swiftascalablelightweightinfrastructure} with DeepSpeed \citep{aminabadi2022deepspeedinferenceenablingefficient} on 8x H100 GPUs. The detailed hyperparameters are provided in Table~\ref{tab:training_hyperparameters_combined}.

\begin{table}[ht]
\centering
\renewcommand{\arraystretch}{1.2}
\begin{tabular}{lcc}
\toprule
\textbf{Parameter} & \textbf{OTG-1.6 (72B)} & \textbf{OTG-R1 (32B)} \\
\midrule
Base Model & \texttt{Qwen2.5-72B-Instruct} & \texttt{DeepSeek-R1-Distill-Qwen-32B} \\
Learning Rate & \multicolumn{2}{c}{$1 \times 10^{-4}$} \\
LoRA Rank & \multicolumn{2}{c}{64} \\
LoRA Alpha & \multicolumn{2}{c}{128} \\
Epochs & \multicolumn{2}{c}{3} \\
GPUs & \multicolumn{2}{c}{8x H100} \\
Training Framework & \multicolumn{2}{c}{MS Swift + DeepSpeed} \\
\midrule
Batch Size per Device & 4 & 2 \\
Gradient Accumulation & 1 & 2 \\
Maximum Length (Initial Training) & 2400 tokens & 8192 tokens \\
Maximum Length (Extended Training) & N/A & 16384 tokens \\
\bottomrule
\end{tabular}
\vspace{0.3cm}
\caption{Training Hyperparameters for OTG-1.6 (72B) and OTG-R1 (32B). Common hyperparameters are displayed in unified rows.}
\label{tab:training_hyperparameters_combined}
\end{table}

\subsection{OTG-1.6: General Model (72B)}

OTG-1.6 was developed to achieve broad generalization across various domains. The model training involved multiple datasets covering general instructions, translation pairs, and standardized Thai examinations. The general instruction dataset was derived from \texttt{Thaweewat/alpaca-cleaned-52k-th}\footnote{\url{https://huggingface.co/datasets/Thaweewat/alpaca-cleaned-52k-th}}, \texttt{OpenAssistant/oasst1}\footnote{\url{https://huggingface.co/datasets/OpenAssistant/oasst1}}, and \texttt{Thaweewat/gpteacher-20k-th}\footnote{\url{https://huggingface.co/datasets/Thaweewat/gpteacher-20k-th}}. Bilingual datasets from Lexitron (English and Thai) were included to enhance cross-lingual understanding. Additionally, standardized examination datasets, including ONET and TGAT, were incorporated to improve domain-specific performance.

\textbf{Model Merging:} The models trained on these datasets were subsequently merged using \texttt{mergekit}~\citep{goddard-etal-2024-arcees} through Task Arithmetic model merging. This process involved weighted merging of specialized models to enhance generalization without increasing computational requirements. The merging weights were assigned as follows: \textbf{General Instructions Model (0.15)}, \textbf{Translation Pairs Model (0.15)}, and \textbf{Thai Exams Model (0.70)}. Model merging facilitated the integration of specialized knowledge from distinct datasets, enhancing the model's robustness and performance across diverse tasks.

\subsection{OTG-R1: Reasoning Model (32B)}

OTG-R1 was developed to enhance reasoning capabilities, particularly for structured and complex tasks. The model was initially trained using various instruction-based datasets, including \texttt{Thaweewat/alpaca-cleaned-52k-th}\footnote{\url{https://huggingface.co/datasets/Thaweewat/alpaca-cleaned-52k-th}}, \texttt{OpenAssistant/oasst1}\footnote{\url{https://huggingface.co/datasets/OpenAssistant/oasst1}}, \texttt{Thaweewat/gpteacher-20k-th}\footnote{\url{https://huggingface.co/datasets/Thaweewat/gpteacher-20k-th}}, \texttt{iapp/Thai-R1-Distill-SFT}\footnote{\url{https://huggingface.co/datasets/iapp/Thai-R1-Distill-SFT}}, and \texttt{ServiceNow-AI/R1-Distill-SFT}\footnote{\url{https://huggingface.co/datasets/ServiceNow-AI/R1-Distill-SFT}}. Additionally, standardized examination datasets, including ONET and TGAT, were incorporated to improve foundational reasoning capabilities through comprehensive dataset coverage.

\textbf{Multi-Stage Training:} OTG-R1 employed a progressive training process involving multiple rounds to refine reasoning capabilities. The initial training phase utilized a maximum sequence length of \textbf{8192 tokens}, which was extended to \textbf{16384 tokens} in the subsequent phase. This progression allowed the model to process increasingly complex queries while maintaining computational efficiency.

\textbf{LIMO Integration:} To enhance reasoning performance with limited high-quality data, the LIMO dataset\footnote{\url{https://github.com/GAIR-NLP/LIMO}} was incorporated during the second training phase. This dataset was applied without translation to preserve contextual integrity. The integration of LIMO aimed to leverage high-quality, contextually relevant data to improve reasoning efficiency and accuracy.

\section{Experiments}

The trained models were evaluated on multiple benchmarks that cover a diverse range of tasks, including mathematical reasoning, language comprehension, coding skills, and general knowledge. The evaluation aimed to assess the models' generalization, reasoning capabilities, and consistency across tasks.

\subsection{\textbf{Benchmarks}}

The evaluation used various benchmarks designed to test the specific abilities of the models. Each benchmark is described in the following:

\begin{itemize}
    \item \textbf{AIME24-TH}: A Thai translation of the American Invitational Mathematics Examination (AIME) dataset focusing on advanced mathematical reasoning and problem solving.
    \item \textbf{MATH500-TH}: A Thai-specific mathematical reasoning dataset curated to evaluate calculation skills, logic, and conceptual understanding.
    \item \textbf{LiveCodeBench-TH}: A coding benchmark adapted for Thai that evaluates the models' ability to understand, generate, and debug code snippets based on Thai-language prompts.
    \item \textbf{OpenThaiEval}: A comprehensive evaluation suite designed for Thai language models that assessesses general knowledge, comprehension, and contextual reasoning.
    \item \textbf{Language Accuracy}: A benchmark aimed at evaluating the models' consistency and accuracy in generating responses in Thai, particularly focusing on grammatical correctness and coherence.
\end{itemize}

\subsection{Results}

The performance of OTG-1.6 and OTG-R1 was evaluated against several existing models. The results are summarized in Tables~\ref{tab:llm_benchmark} and \ref{tab:llm_reasoning_benchmark}.

\subsubsection{General Model}

OTG-1.6 was compared to previous OpenThaiGPT versions and other Thai LLMs, including Typhoon2 and Pathumma. The evaluation demonstrates substantial improvements in generalization, particularly on benchmarks such as \textbf{OpenThaiEval} and \textbf{Language Accuracy}. OTG-1.6 consistently outperforms OpenThaiGPT 1.5 models and most Typhoon2 variants on various tasks. Despite maintaining the same model size as previous OpenThaiGPT models, OTG-1.6 achieves superior performance due to Task Arithmetic model merging, which integrates specialized knowledge from diverse datasets without increasing model scale. This efficiency allows OTG-1.6 to effectively balance computational resources and performance.

The model demonstrates state-of-the-art performance on OpenThaiEval (78.70). Furthermore, OTG-1.6 exhibits robust generalization in mathematical reasoning and coding tasks, significantly outperforming previous OpenThaiGPT models and most Typhoon2 variants. The average score of OTG-1.6 across all benchmarks is 52.34, substantially higher than any previous OpenThaiGPT models.

\begin{table}[ht!]
\centering
\renewcommand{\arraystretch}{1.1}
\begin{tabular}{lcccccccc}
\toprule
\multirow{2}{*}{\textbf{Benchmarks}} & \textbf{OTG-1.6} & \multicolumn{3}{c}{\textbf{OTG-1.5}} & \multicolumn{3}{c}{\textbf{Typhoon2}} & \textbf{Pathumma} \\
\cmidrule(lr){2-2} \cmidrule(lr){3-5} \cmidrule(lr){6-8} \cmidrule(lr){9-9}
 & 72b \textdagger & 7b \textdagger & 14b \textdagger & 72b \textdagger & 7b \textdagger & 8b \textsection & 70b \textsection & 1.0.0 \\
\midrule
AIME24-TH         & 6.67   & 0.00   & 0.00   & 6.67   & 3.33   & 3.33   & \textbf{13.33} & 0.00 \\
AIME24            & \textbf{23.33}  & 6.67  & 10.00  & \textbf{23.33}   & 6.67   & 3.33   & 10.00  & 0.00 \\
MATH500-TH        & 43.20  & 24.20  & 26.20  & \textbf{62.00}  & 51.80  & 31.00  & 55.80  & 21.80 \\
MATH500           & 82.00  & 40.40  & 47.40  & \textbf{83.20}  & 65.40  & 49.60  & 67.40  & 42.80 \\
LiveCodeBench-TH  & \textbf{32.43}  & 22.52  & 21.62  & 12.61  & 9.91   & 8.11   & 27.03  & 0.00 \\
LiveCodeBench     & \textbf{54.21}  & 31.12  & 37.96  & 46.38  & 0.98   & 5.87   & 37.38  & 0.00 \\
OpenThaiEval      & \textbf{78.70}  & 64.50  & 71.26  & 77.16  & 64.76  & 56.63  & 72.54  & 65.27 \\
Language Accuracy & 98.20  & 97.60  & 98.40  & 99.40  & 99.40  & 98.60  & \textbf{99.80}  & 98.60 \\
\midrule
\textbf{Average}  & \textbf{52.34}  & 35.88  & 39.11  & 51.34  & 37.78  & 32.06  & 47.91  & 28.56 \\
\bottomrule
\end{tabular}
\vspace{0.3cm}
\caption{Benchmark Results Across Multiple Models. \textdagger~indicates Qwen2.5 architecture and \textsection~indicates Llama3.1 architecture.}
\label{tab:llm_benchmark}
\end{table}

\subsubsection{Reasoning Model}

OTG-R1 was evaluated specifically for reasoning tasks against baseline models, including DeepSeek-R1 and Typhoon2-R1. Despite having a smaller model size (32B) compared to the 70B baselines, OTG-R1 consistently achieves superior or competitive results across various reasoning benchmarks. This highlights the effectiveness of the LIMO integration and multi-stage training techniques applied during model development.

OTG-R1 achieves the highest scores on critical reasoning benchmarks, including AIME24-TH (56.67) and MATH500-TH (83.80). Additionally, it demonstrates strong performance on LiveCodeBench-TH (62.16) and LiveCodeBench (69.67), outperforming DeepSeek-R1 and Typhoon2-R1 despite having a significantly smaller model size. The overall average score of OTG-R1 is 71.59, surpassing both DeepSeek-R1 (63.32) and Typhoon2-R1 (65.43). These results indicate that targeted training methodologies are more effective than scaling alone, particularly when combined with structured fine-tuning techniques like LIMO.

\begin{table}[ht!]
\centering
\renewcommand{\arraystretch}{1.1}
\begin{tabular}{lccc}
\toprule
\multirow{2}{*}{\textbf{Benchmarks}} & \textbf{OTG-R1} & \textbf{DeepSeek-R1} & \textbf{Typhoon2-R1} \\
\cmidrule(lr){2-2} \cmidrule(lr){3-3} \cmidrule(lr){4-4}
 & 32b & 70b & 70b \\
\midrule
AIME24-TH         & \textbf{56.67}   & 33.33  & 53.33 \\
AIME24            & \textbf{63.36}   & 53.33  & 53.33 \\
MATH500-TH        & \textbf{83.80}   & 75.40  & 81.00  \\
MATH500           & 89.40            & 88.88  & \textbf{90.20} \\
LiveCodeBench-TH  & \textbf{62.16}   & 53.15  & 47.75 \\
LiveCodeBench     & \textbf{69.67}   & 64.97  & 54.79 \\
OpenThaiEval      & 76.05            & 74.17  & \textbf{77.59} \\
\midrule
\textbf{Average}  & \textbf{71.59}   & 63.32  & 65.43 \\
\bottomrule
\end{tabular}
\vspace{0.3cm}
\caption{Reasoning Benchmark Results Across Multiple Models.}
\label{tab:llm_reasoning_benchmark}
\end{table}

\section{Discussion}
The experimental results confirm the effectiveness of the proposed training methodologies for both OTG-1.6 and OTG-R1. OTG-1.6 demonstrates strong generalization capabilities through Task Arithmetic model merging. Despite maintaining the same scale as previous models, OTG-1.6 achieves superior performance across various benchmarks, including OpenThaiEval and Language Accuracy. The integration of specialized models enhances generalization and robustness without increasing computational requirements. This approach offers a scalable solution for broader generalization tasks while preserving efficiency.

OTG-R1, optimized for reasoning through multi-stage training and LIMO integration, achieves competitive or superior results on various reasoning benchmarks. Remarkably, it surpasses larger models such as DeepSeek-R1 and Typhoon2-R1 despite its smaller size (32B compared to 70B). This highlights the efficiency of combining LIMO with progressive training techniques. The performance of OTG-R1 demonstrates that a well-structured training process can compensate for reduced model size, offering an efficient alternative to conventional scaling.

The results suggest that Task Arithmetic model merging and reasoning-specific training are effective strategies for enhancing performance on Thai-centric benchmarks. While OTG-1.6 excels in generalization, OTG-R1 demonstrates the benefits of reasoning-specific optimization. The complementary strengths of these models provide valuable insights into developing both specialized and general-purpose Thai-centric LLMs.

\section{Conclusion}

This paper presents OTG-1.6 and OTG-R1, two Thai-centric LLMs optimized for generalization and reasoning. OTG-1.6 employs Task Arithmetic model merging to enhance knowledge representation across diverse domains without increasing model size. OTG-R1 utilizes multi-stage training and LIMO integration to achieve strong reasoning performance, surpassing larger models such as DeepSeek-R1 and Typhoon2-R1 despite its smaller size. Both models demonstrate competitive results across various benchmarks, outperforming previous models and most baseline LLMs.

The results demonstrate the effectiveness of specialized training techniques, including Task Arithmetic model merging, multi-stage training, and LIMO integration. These approaches enable efficient knowledge integration and improved reasoning capabilities. Future work will focus on expanding evaluation benchmarks, enhancing training methodologies, and optimizing models for broader applications.

\section*{Limitations}

\textbf{Model Merging Efficiency}:
The model merging technique used in OTG-1.6 enhances generalization but may overlook domain-specific knowledge due to weight distribution limitations.

\textbf{Computational Demands}:
OTG-R1's fine-tuning process requires significant computational resources, especially during multi-stage training with extended sequence lengths.

\textbf{Dataset Dependency}:
The effectiveness of LIMO integration relies heavily on the quality and diversity of available datasets. Limited access to high-quality Thai-specific datasets may affect performance.

\textbf{Benchmark Coverage}:
Current evaluation benchmarks do not cover all aspects of Thai language understanding and reasoning, potentially overlooking certain capabilities.

\textbf{Out-of-Domain Weaknesses}:
The models may struggle with out-of-domain inputs or highly specialized queries that were not adequately addressed during training.

Addressing these limitations will be essential for further improving Thai-centric LLMs, particularly in domain-specific applications and real-world scenarios.

% \clearpage  % Forces a new page

\appendix
\section{Project Links}

The following are the official links for the OpenThaiGPT project and model releases:

\begin{itemize}
    \item \textbf{Official Project Website:} \url{https://openthaigpt.aieat.or.th/}
    \item \textbf{OpenThaiGPT 1.6 72B:} \url{https://huggingface.co/openthaigpt/openthaigpt-1.6-72b-instruct}
    \item \textbf{OpenThaiGPT R1 32B:} \url{https://huggingface.co/openthaigpt/openthaigpt-r1-32b-instruct}
\end{itemize} 

\bibliographystyle{unsrtnat}
\bibliography{references}  %%% Uncomment this line and comment out the ``thebibliography'' section below to use the external .bib file (using bibtex) .




% 	\bibitem{keshet2016prediction}
% 	Keshet, Renato, Alina Maor, and George Kour.
% 	\newblock Prediction-Based, Prioritized Market-Share Insight Extraction.
% 	\newblock In {\em Advanced Data Mining and Applications (ADMA), 2016 12th International 
%                       Conference of}, pages 81--94,2016.

% \end{thebibliography}

\end{document}